\def\year{2022}\relax
\documentclass[letterpaper]{article} % DO NOT CHANGE THIS
\usepackage{aaai22}  % DO NOT CHANGE THIS
\usepackage{times}  % DO NOT CHANGE THIS
\usepackage{helvet}  % DO NOT CHANGE THIS
\usepackage{courier}  % DO NOT CHANGE THIS
\usepackage[hyphens]{url}  % DO NOT CHANGE THIS
\usepackage{graphicx} % DO NOT CHANGE THIS
\usepackage{natbib}  % DO NOT CHANGE THIS AND DO NOT ADD ANY OPTIONS TO IT
\usepackage{caption} % DO NOT CHANGE THIS AND DO NOT ADD ANY OPTIONS TO IT
\DeclareCaptionStyle{ruled}{labelfont=normalfont,labelsep=colon,strut=off} % DO NOT CHANGE THIS
\usepackage[linesnumbered,ruled,vlined]{algorithm2e}
\usepackage{algpseudocode}  

\usepackage{amsmath}  
\usepackage{graphicx}
\usepackage{caption}
\usepackage{natbib}
\usepackage{amsthm,amsmath,amssymb}
\usepackage{mathrsfs}
\usepackage{soul}
\usepackage{url}
\usepackage{enumerate}
\usepackage[utf8]{inputenc}
\usepackage{mathtools}
\usepackage{subfigure}   
\usepackage{cancel}
\usepackage{comment}
\usepackage{ctable}
\usepackage{mathtools}
\usepackage{balance}
\usepackage{threeparttable}  
\usepackage{booktabs}
\usepackage{multirow}
\usepackage{bm}
\usepackage{enumitem}
\usepackage[switch]{lineno} 
\usepackage{xcolor,colortbl}
\theoremstyle{definition}   
\newtheorem{Definition}{Definition}

\usepackage{newfloat}
\usepackage{listings}
\title{Instance-wise Graph-based Framework for Multivariate Time Series Forecasting}

\author{
	Wentao Xu,\textsuperscript{\rm 1,4}\thanks{Work done during an internship at Microsoft Research Asia.}
	Weiqing Liu, \textsuperscript{\rm 2}
	Jiang Bian, \textsuperscript{\rm 2}
	Jian Yin, \textsuperscript{\rm 3,4}
	Tie-Yan Liu \textsuperscript{\rm 2}
}
\affiliations{
	\textsuperscript{\rm 1}School of Computer Science and Engineering, Sun Yat-sen University, Guangzhou, China\\
	\textsuperscript{\rm 2}Microsoft Research Asia, Beijing, China\\
	\textsuperscript{\rm 3}School of Artificial Intelligence, Sun Yat-sen University, Zhuhai, China\\
	\textsuperscript{\rm 4}Guangdong Key Laboratory of Big Data Analysis and Processing, Guangzhou, China\\
	\{xuwt6@mail2,issjyin@mail\}.sysu.edu.cn\\
	\{weiqing.liu,jiang.bian,tyliu\}@microsoft.com
}
\usepackage{bibentry}
\begin{document}

\maketitle

\begin{abstract}
The multivariate time series forecasting has attracted more and more attention because of its vital role in different fields in the real world, such as finance, traffic, and weather. In recent years, many research efforts have been proposed for forecasting multivariate time series. Although some previous work considers the interdependencies among different variables in the same timestamp, existing work overlooks the inter-connections between different variables at different time stamps. In this paper, we propose a simple yet efficient instance-wise graph-based framework to utilize the inter-dependencies of different variables at different time stamps for multivariate time series forecasting. The key idea of our framework is aggregating information from the historical time series of different variables to the current time series that we need to forecast. We conduct experiments on the Traffic, Electricity, and Exchange-Rate multivariate time series datasets. The results show that our proposed model outperforms the state-of-the-art baseline methods.
\end{abstract}

\maketitle

\section{Introduction}
\label{sec:intro}
Multivariate time series have more than one time-dependent variable. Each variable depends on its historical values as well as other variables.
Multivariate time series exist in many aspects of our daily lives, including the price series in the stock market, the occupancy rates of different roads, the temperatures and rainfalls across various cities.
Mining the meaningful information from multivariate time series and forecasting the time series' future trends can benefit many domains of human society, such as finance, traffic governance, and weather forecasting.

In the past decades, many research efforts have investigated the multivariate time series forecasting problems. 
Traditional statistical methods like auto-regressive model (AR), ARIMA model~\cite{brown2004smoothing} and Gaussian process model (GP)~\cite{roberts2013gaussian} assume a linear dependency among variables. Thus their model complexity grows quadratically with the number of variables and has the problem of overfitting with a large number of variables~\cite{wu2020connecting}.
With the development of deep learning, some deep learning based methods, including LSTM~\cite{hochreiter1997long}, GRU~\cite{chung2014empirical}, LSTNet~\cite{lai2018modeling}, and TPA-LSTM~\cite{shih2019temporal}, utilize the deep neural network to capture the non-linear patterns of multivariate time series.
More recently, to exploit the latent inter-dependencies among different variables in a multivariate time series, the MTGNN~\cite{wu2020connecting} and StemGNN~\cite{cao2021spectral} use the variables as nodes to construct a graph, and leverage the graph neural networks (GNNs)~\cite{kipf2017semi} to mine the interactions among variables in the same timestamp.

\begin{figure}[t]
	\centering
	\includegraphics[width=0.47\textwidth]{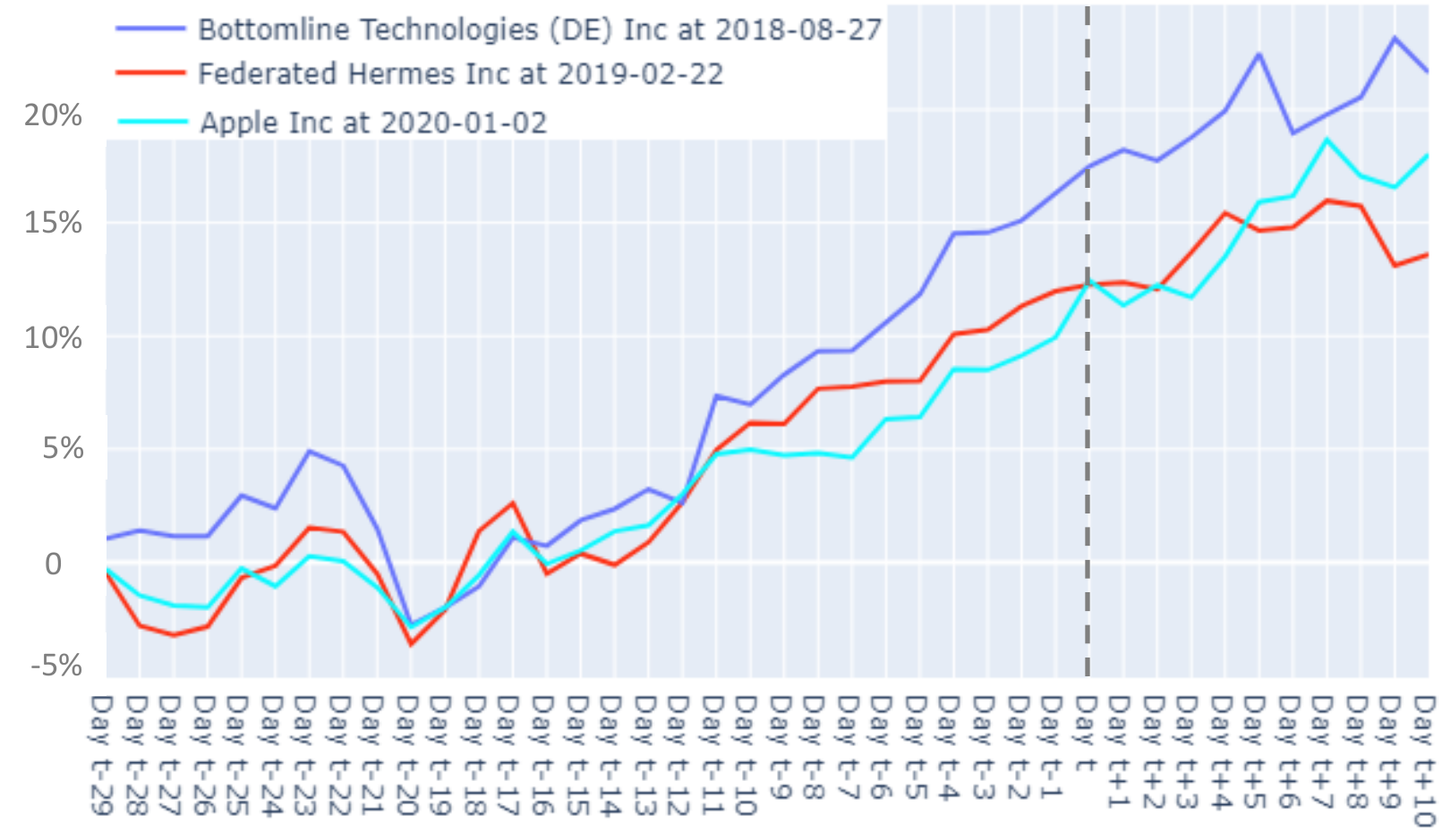}
	\caption{The cumulative return series of three stocks at different time stamps. Day t is the current time stamp (2019-02-02 for Federated Hermes Inc; 2020-01-02 for Apple Inc). Day t - 29 to Day t is the past 30 days, and Day t + 1 to Day t + 10 is the future 10 days.}
	\label{fig:example}
\end{figure}   

Although some previous work considers the interdependencies among different variables in the same timestamp, they overlook the inter-connections between a variable and other variables' historical series. 
Due to the similar external environment or the periodicity of time series, a variable would have a similar series as other variables' historical series.
Figure~\ref{fig:example} shows the cumulative return series of three stocks in different timestamps. 
Although there are not at identical timestamps, they still have similar time series, which implies that different variables at different timestamps would also have inter-dependencies. 
The inter-dependencies between historical time series and the current time series that we need to forecast are valuable.
We can utilize these inter-dependencies and the historical time series of different variables to improve the multivariate time series forecasting.

In this paper, we propose a simple yet efficient instance-wise graph-based framework for multivariate time series forecasting (IGMTF), which utilizes the inter-dependencies between different variables at different time stamps.
We first introduce the concept of series instance, which represents the observed values of a variable at a time stamp. 
Each series instance has a specific input series as the feature. 
Then we have the training instances in the training set and the mini-batch instances in each training/inference mini-batch.
In general, the dataset's training/validation/test sets are split by chronological order, so the training instances contain the historical series of different variables.
We first use the training instance encoder and mini-batch instance encoder to encode the training and mini-batch instances.
Then we utilize a training instances sampler to sample the most related training instances for each mini-batch as sample training instances.
To capture the inter-dependencies between the historical series of different variables and the current series in the mini-batch, we construct an instance graph with the nodes of sampled training instances and mini-batch instances. The edges in instance graph are the connections between these two types of instances.
Then we aggregate information from sampled training instances to mini-batch instances on the instance graph. Thus we can capture the inter-dependencies between the historical series of variables and the current series in mini-batch.
Finally, we utilize aggregated information on mini-batch instances and the mini-batch instances' information to forecast the time series.

We evaluate our framework on three multivariate time series benchmarks datasets: Traffic, Electricity, and Exchange-Rate. The experimental results show that our method outperforms existing multivariate time series methods. Moreover, the empirical analyses verify the effectiveness of some components and influence of some hyper-parameters in our IGMTF framework.

\section{Related Work}
\label{sec:related}
In recent years, there have been many research efforts on multivariate time series forecasting problems.
Some traditional linear regression methods include auto-regressive (AR), vector auto-regressive (VAR)~\cite{zhang2003time}, auto-regressive moving average (ARMA), auto-regressive integrated moving average (ARIMA)~\cite{brown2004smoothing}, and support vector regression (SVR)~\cite{cao2003support}. 
They utilize the linear function of past time historical values to forecast the time series. 
Gaussian process (GP)~\cite{roberts2013gaussian} is a Bayesian approach, modeling the distribution of a multivariate variable over functions, and can naturally apply to model multivariate time series data.
Although traditional linear statistical models have the advantages of simplicity and interpretability, they have the limitation of strong assumptions with respect to a stationary process, and they do not scale well to multivariate time series data.

More recently, more and more deep learning based methods have been proposed because they are free from stationary assumptions and can capture the non-linearity patterns of series~\cite{bai2018empirical,sen2019think,guo2019attention}. 
One representative type of method is the recurrent neural networks (RNNs) and its variants such as long short-term memory (LSTM)~\cite{hochreiter1997long} and gated recurrent units (GRU)~\cite{chung2014empirical}.    
Some RNNs are specially designed for multivariate time series forecasting problems, such as LSTNet~\cite{lai2018modeling} and TPA-LSTM~\cite{shih2019temporal}. The LSTNet utilizes convolutional neural networks (CNNs) to capture local dependencies among variables and RNNs to preserve the long-term temporal dependencies.
The TPA-LSTM utilizes the attention mechanism to capture the long-term dependencies of multivariate time series.
Nevertheless, the LSTNet and TPA-LSTM cannot fully exploit latent dependencies between pairs of variables. 
To capture the inter-dependencies among different variables in multivariate time series, MTGNN~\cite{wu2020connecting} and StemGNN~\cite{cao2021spectral} use the variables as nodes to construct a graph, and leverage graph neural networks (GNNs)~\cite{kipf2017semi} to mine the correlations among variables in the same time stamp.

However, previous multivariate time series forecasting methods overlook the inter-connections between different variables at different time stamps.
Therefore, we propose an instance-wise graph-based framework to utilize the inter-dependencies between variables at different time stamps to boost multivariate time series forecasting.

Besides, some methods focus on the univariate time series forecasting problems~\cite{patel2015predicting,rather2015recurrent,xingjian2015convolutional,zhang2017stock,oreshkin2019n,montero2020fforma}. 
The main difference between univariate and multivariate time series forecasting methods is that univariate time series techniques analyze each time series separately without considering the correlations between different variables. Since considering correlations between different variables at different time series is the critical insight of our work, we do not introduce the univariate time series techniques in detail.
\begin{figure*}[t]
	\centering
	\includegraphics[width=0.88\textwidth]{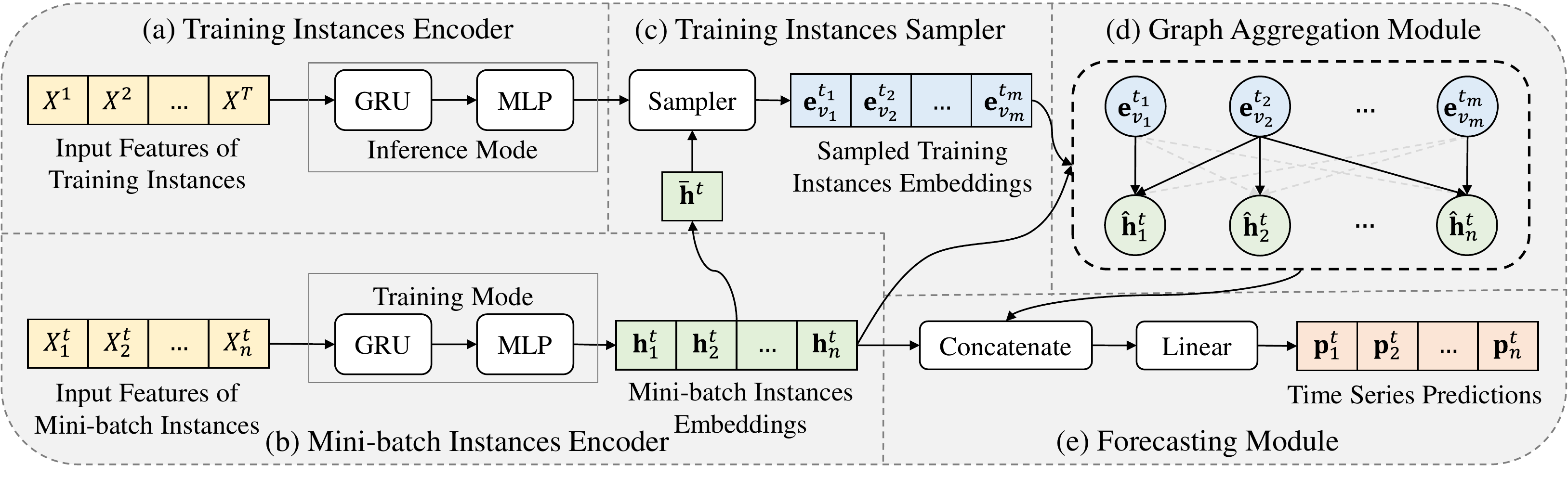}
	\caption{The overall architecture of the proposed IGMTF framework.}
	\label{fig:framework}
\end{figure*}
\section{Preliminaries}
\label{subsec:preliminaries}
In this section, we first formulate the problem of multivariate time series forecasting.
Give a sequence observations on a multivariate variable at time stamp $t$: $X^t = \{\mathbf{x}^{t-d}, \mathbf{x}^{t-d+1}, ..., \mathbf{x}^t\}$, where $d$ is the length of input time stamps, $\mathbf{x}^{t-k} \in \mathbb{R}^n$ and $n$ is the dimension of variables, our goal is to predict the future series value $\mathbf{x}^{t+h}$, where $h$ is the specified horizon ahead of the current timestamp.
The horizon $h$ can be set to specific values according to the type of time series data.
The multivariate time series forecasting is a rolling process. At timestamp $t$, when we forecast the future value $\mathbf{x}^{t+h}$, we assume the sequence $\{\mathbf{x}^{t-d}, \mathbf{x}^{t-d+1}, ..., \mathbf{x}^t\}$ is available. Similarly, when we forecast the value $\mathbf{x}^{t+h+1}$ at timestamp $t+1$, we assume the sequence $\{\mathbf{x}^{t-d+1}, \mathbf{x}^{t-d+2}, ..., \mathbf{x}^{t+1}\}$ is available. 

Besides, we would also describe the formal definition of some concepts in our framework below.

\begin{Definition}
	\textbf{Series Instance}. A series instance $v_i^t$ is the observed values of a variable $v_i$ at timestamp $t$.
	The feature of series instance $v_i^t$ is $X_i^t = \{\mathbf{x}_i^{t-d}, \mathbf{x}_i^{t-d+1}, ..., \mathbf{x}_i^t\}$, where $X_i^t\in \mathbb{R}^d$, and the forecasting label of series instance $v_i^t$ is $\mathbf{x}_i^{t+h}$. 
	In this paper, we also call the series instance as instance for convenience.
\end{Definition}
In a multivariate time series, each variable at each timestamp would have a series instance. If a time series has 30 variables at 100 timestamps, then the number of series instances in this time series is 3000.
\begin{Definition}
	\textbf{Instance Graph}. The instance graph is a graph whose nodes are the series instances. The edges are the similarity between different series instances.
\end{Definition}

\section{Our Framework}
\label{sec:method}
 
This section elaborates on our proposed instance-wise graph-based framework for multivariate time series forecasting (IGMTF). Figure~\ref{fig:framework} shows the overall architecture of our framework. 
The key idea of our framework is capturing the interdependencies between historical instances of different variables and the current instances in the mini-batch.
To achieve this goal, in Section~\ref{subsec:training-encoder} and~\ref{subsec:mini-batch-encoder}, we first learn the embeddings of training and mini-batch instances. 
In Section~\ref{subsec:sampler}, we use a training instance sampler to sample the most related training instances for each mini-batch. After that, in Section~\ref{subsec:graph}, we utilize the graph aggregation module to aggregate information from sampled training instances to mini-batch instances.
Finally, in the forecasting module in Section~\ref{subsec:forecast}, we use the aggregated information on mini-batch instances as well as the mini-batch instances' embeddings to forecast the future time series.
\subsection{Training Instances Encoder}
\label{subsec:training-encoder}
The training instances encoder aims to learn the representations of series instances in training set $D_{train} = \{X^1, X^2, ..., X^T\}$, where $X^i\in \mathbb{R}^{n\times d}$ and $T$ is the number of time stamps in the training set. 
At the beginning of each training epoch or model inference, we encode the features of all instances in $D_{train}$.
Since the feature of each instance is the historical values of a variable, we need to encode the historical information of each instance.
The Gated Recurrent Unit (GRU) network~\cite{chung2014empirical} can capture the long-term dependency of series, so we feed the features of training instances in $D_{train}$ into a GRU.
Then we further project the last hidden state of GRU's output with a 3-layers MLP, and the output of MLP is the training instance embeddings $E =\{E^1, E^2, ..., E^T\}$, where $E^i \in \mathbb{R}^{n \times l}$ and $l$ is the dimension of MLP's output.
In this MLP, there is a LeakyReLU~\cite{maas2013rectifier} activation function after each linear layer.
It is noteworthy that we feed the training instances into the GRU and MLP in inference mode, without generating any gradients in this process.

\subsection{Mini-batch Instances Encoder}
\label{subsec:mini-batch-encoder}
The mini-batch instances encoder aims to learn the representations of series instances in a mini-batch $\mathcal{D}_{batch}$.
In our framework, each mini-batch is all the instances at the same time stamp, so the number of instances in a mini-batch equals the number of variables $n$ in the dataset.
In each mini-batch $\mathcal{D}_{batch}$, we feed the features $X^t = \{X_1^t, X_2^t,, ..., X_n^t\}$ of $n$ instances $\{v_1^t, v_2^t, ..., v_n^t\}$ at time stamp $t$ into the same GRU in training instances encoder.
We also feed the last hidden layer of GRU's output to the same MLP in training instances encoder to make a projection.
In the mini-batch instance encoder, the GRU and MLP are in training mode. They would generate gradients to update the parameters in GRU and MLP.
The output of MLP $H^t = \{\mathbf{h}_1^t, \mathbf{h}_2^t,, ..., \mathbf{h}_n^t\}$, where $\mathbf{h}_i^t \in  \mathbb{R}^{l}$, is the mibi-batch instance embeddings.

\subsection{Training Instances Sampler}
\label{subsec:sampler}
Since the number of training instances is huge and there is a vast computation cost to directly aggregate information from training instances to mini-batch instances, we utilize a training instances sampler to sample the most related training instances $\mathcal{D}_{train}^{sample}$ from $\mathcal{D}_{train}$ for each mini-batch.
We first calculate the mean value of training instance embeddings at each time stamp: $\bar{E} = \{\bar{E}^1, \bar{E}^2, ..., \bar{E}^T\}$, and the mean value mini-batch instance embeddings $H^t$.
\begin{equation} 
	\bar{E}^i  = \dfrac{1}{n} \sum_{j=1}^n E^i_j,
\end{equation}
\begin{equation}
	\bar{\mathbf{h}}^t  = \dfrac{1}{n} \sum_{j=1}^n \mathbf{h}^t_j,
\end{equation}
where $\bar{E}^i \in \mathbb{R}^l$ is the mean embedding of training instances at time stamp $i$, and $\bar{\mathbf{h}}^t  \in \mathbb{R}^l$ is the mean embedding of mini-batch instance embeddings $H^t$.

To find the most related instances for each mini-batch, we calculate the cosine similarity between each time stamp's embedding $\bar{E}^i$ and $\bar{\mathbf{h}}^t$. We select the closest $k$ time stamp embeddings with $\bar{\mathbf{h}}^t$ according to the similarities.
After that, we sample the training instances in these $k$ time stamps as sampled training time stamps. Since each time stamp has $n$ training instances, the total number of sampled training instances $\mathcal{D}_{train}^{sample}$ is $m = n \times k$.
We would study the effect of training instances sampler in Section~\ref{subsec:ablation}.

\subsection{Graph Aggregation Module}
\label{subsec:graph}
In the graph aggregation module, we capture the inter-dependencies between the sampled training instances $\mathcal{D}_{train}^{sample}$ and the mini-batch instances $\mathcal{D}_{batch}$.
We first use the instances in $\mathcal{D}_{train}^{sample}$ and $\mathcal{D}_{batch}$ as nodes to construct a instance graph. In this instance graph, each sampled training instance and mini-batch instance are connected with an edge.
Given $m$ sampled training instance embeddings $E_s=\{\mathbf{e}_{v_1}^{t_1}, \mathbf{e}_{v_2}^{t_2}, ..., \mathbf{e}_{v_m}^{t_m}\}$ and $n$ mini-batch instance embeddings  $H^t = \{\mathbf{h}_1^t, \mathbf{h}_2^t, ...,\mathbf{h}_n^t\}$, we aggregate the information from sampled training instances to mini-batch instances on the instance graph.
Since the instance graph is not pre-defined and we do not know the weight between sampled training instances and mini-batch instances, we use the cosine similarity between sampled training instance embeddings $E_s$ and mini-batch instance embeddings $H^t$ as aggregated weights.
\begin{equation}
	A_{ij} = \mathrm{Cosine} \left(W_h \mathbf{h}_i^t, W_e \mathbf{e}_{v_j}^{t_j}\right)= \dfrac{W_h \mathbf{h}_i^t \cdot W_e \mathbf{e}_{v_j}^{t_j}}{|| W_h \mathbf{h}_i^t || \cdot || W_e \mathbf{e}_{v_j}^{t_j}||},
\end{equation}
where $W_h$ and $W_e$ are two mapping matrices for $\mathbf{h}_i^t$ and $\mathbf{e}_{v_j}^{t_j}$, respectively. We would study the effect of these two mapping matrices in Section~\ref{subsec:ablation}.
The $A_{ij}$ is the similarity between mini-batch instance embedding $\mathbf{h}_i^t$ and sampled training instance embedding $\mathbf{e}_{v_j}^{t_j}$, the matrix $A$ is the adjacent matrix from sampled training instances to mini-batch instances.

The current adjacent matrix $A$ indicates a fully connected graph since each training instance would calculate the similarity with each mini-batch instance.
Some recent work~\cite{over-smoothing, liu2020towards} have pointed out that the deeper graph neural network (GNN) would cause the problem of over-smoothing, which is the repeated message propagation makes nodes in different classes have indistinguishable representations. 
Similarly, the GNN on a fully connected graph would also induce the issue of over-smoothing.
Therefore, we introduce a top $N$ mask mechanism on the adjacent matrix $A$.
For the aggregated weights $A_{ij}$, where $j \in [1, m]$, from all training instances to the mini-batch instance $v_i$, we only retain the top $N$ largest weights and mask the remaining weights as $0$.
In this way, we only retain the closest $N$ neighbors for each mini-batch instance.
Then we utilize the masked adjacent matrix $\hat{A}$ to aggregate information from the most closest $N$ training instances to each mini-batch instance:
\begin{equation}
	\hat{\mathbf{h}}_i^t =\dfrac{1}{|\mathcal{N}_i|} \sum_{j\in \mathcal{N}_i} \hat{A}_{ij} W_e \mathbf{e}_{v_j}^{t_j},
\end{equation}
where $\mathcal{N}_i$ is the set of top $N$ closest training instances for the mini-batch instance $v_i^t$.
The $\hat{\mathbf{h}}_i^t$ is the information aggregation from the historical training instances to the instance $v_i^t$ at time stamp $t$.

\subsection{Forecasting Module}
\label{subsec:forecast}
Finally, we combine aggregated information and mini-batch instances embedding to forecast the future time series jointly.
We concatenate aggregation information from training instances $\hat{\mathbf{h}}_i^t$ and mini-batch instance embedding $\mathbf{h}_i^t$, and feed the concatenation into a linear layer. The $1$ dimensional output of linear layer is the time series forecasting $\mathbf{p}_i^t$, which is the prediction to the instance $v_i^t$'s label $\mathbf{x}_i^{t+h}$.
\begin{equation}      
\mathbf{p}_i^t = \mathrm{Linear}(\mathrm{Concat}(\hat{\mathbf{h}}_i^t, \mathbf{h}_i^t)) .
\end{equation}
\begin{algorithm}[t]
	\SetAlgoLined
	\KwIn{Training data $\mathcal{D}_{train}$, the initialized  model parameters $\Theta$, learning rate $\gamma$.}
	\KwOut{Time series forecasting $P$.}
	\While{not meet the stopping criteria} {
		Encode the features of training instances in $\mathcal{D}_{train}$ as training instance embeddings $E$\;
		\For{$\mathcal{D}_{batch}$ in $ \mathcal{D}_{train}$} {
			Encoder the features of mini-batch instances in the mini-batch $\mathcal{D}_{batch}$ as embeddings $H^t$\;
			Sample the most related training instances $\mathcal{D}_{train}^{sample}$ from $\mathcal{D}_{train}$ for the $\mathcal{D}_{batch}$\;
			Aggregate information from $\mathcal{D}_{train}^{sample}$ to $\mathcal{D}_{batch}$\;
			Forecast the future time series $\mathbf{p}^t$\;
			Compute the stochastic gradients of $\Theta$ with Equation~\ref{eq:loss}\;
			Update model parameters $\Theta$ according to the gradients and learning rate $\gamma$\;
		}
	}
	\caption{The algorithm of our framework.}
	\label{alg:training}
\end{algorithm}

\subsection{Model Training}
We use stochastic gradient descent algorithm (SGD) with mini-batches to train our framework and leverage the Adam~\cite{kingma:adam} for tuning the learning rate. We optimize our framework by minimizing the MAE loss function with L2 regularization:
\begin{equation}
%	\mathcal{L} = \sum_{\mathcal{D}_{batch} \subset \mathcal{D}_{train} }\left( \dfrac{\sum_{v_i^t \in \mathcal{D}_{batch}} |p_i^t - x_i^{t+h}|}{|\mathcal{D}_{batch}|}+ \lambda ||\Theta||_2^2\right),
\mathcal{L} = \dfrac{\sum_{v_i^t \in \mathcal{D}_{batch}} |\mathbf{p}_i^t - \mathbf{x}_i^{t+h}|}{|\mathcal{D}_{batch}|}+ \lambda ||\Theta||_2^2.
\label{eq:loss}
\end{equation}
where $\mathcal{D}_{batch}$ is the instances in a mini-batch, $\mathbf{p}_i^t$ and  $\mathbf{x}_i^{t+h}$ are future series prediction and label of variable $v_i$ at time stamp $t$. The $\lambda$ is the regularization parameter, and $\Theta$ represents all of the parameters in our framework.
Algorithm~\ref{alg:training} is the training algorithm of our framework.

\section{Experiments}
In this section, we present thorough empirical studies to evaluate and analyze our proposed IGMTF framework.
We first introduce the datasets and experimental setting. Then we compare our IGMTF framework with exiting multivariate time series forecasting methods on the benchmark datasets. Moreover, we apply an ablation study to study the effect of some components in our framework. Finally, we analyze the influence of some parameters on our framework.

\subsection{Datasets} We evaluate our method on the three multivariate time series forecasting benchmark datasets: Traffic, Electricity, and Exchange-Rate. All of these datasets are provided by~\cite{lai2018modeling}. Table~\ref{table:time-dataset} shows the statistics of these three datasets. The detailed introduction of datasets are as follows:
\begin{itemize}
	\item Traffic: the traffic dataset from the California Department of Transportation contains road occupancy rates measured by 862 sensors in San Francisco Bay area freeways during 2015 and 2016.
	%	\item Solar-Energy: the solar-energy dataset from the National Renewable Energy Laboratory contains the solar power output collected from 137 PV plants in Alabama State in 2007.
	\item Electricity:  the electricity dataset from the UCI Machine Learning Repository contains electricity consumption for 321 clients from 2012 to 2014.
	\item Exchange-Rate: the exchange-rate dataset contains the daily exchange rates of eight foreign countries, including Australia, British, Canada, Switzerland, China, Japan, New Zealand, and Singapore, ranging from 1990 to 2016.
\end{itemize}

\begin{table}[t]
	\begin{center}
		
		%			\resizebox{\columnwidth}{!}{%
		\begin{tabular}{l|ccc} 
			\toprule
			Datasets 		& \# Time		& \# Variables	& Sample Rate \\
			\midrule
			\midrule
			traffic 		& 17,544 	& 862 	& 1 hour 	\\
			%			solar-energy			& 52,560	& 137	& 10 minutes \\
			electricity 	& 26,304 	& 321 	& 1 hour 	\\
			exchange-rate		& 7,588		& 8		& 1 day	\\
			\bottomrule
		\end{tabular}%}
	\end{center}
	\caption{Datasets statistics, where \# Time and \# Variables are the number of time stamps and variables in the datasets.}
	\label{table:time-dataset}
\end{table}

Following~\cite{lai2018modeling,wu2020connecting}, we split these three datasets into a training set ($60\%$), validation set ($20\%$), and test set ($60\%$) in chronological order. In these three datasets, the input feature length $d$ of an instance is 168 and the forecasting length is 1. Models are trained independently to forecast the target future step (horizon) 3, 6, 12, and 24.

\subsection{Experimental Setting}
\paragraph{Compared Methods}
We compare our framework with the following multivariate time series forecasting methods:
\begin{itemize}
	\item AR: A traditional auto-regressive model.
	\item VAR-MLP~\cite{zhang2003time}: A hybrid model of the vector auto-regressive model (VAR) and MLP.
	\item GP~\cite{roberts2013gaussian}: A Gaussian Process time series model.
	\item RNN-GRU~\cite{chung2014empirical}: A recurrent neural network with fully connected GRU hidden units.
	\item LSTNet~\cite{lai2018modeling}: A deep neural network that combines convolutional neural networks and recurrent neural networks to capture the long-term and short-term temporal
	patterns.
	\item TPA-LSTM~\cite{shih2019temporal}: An attention-based recurrent neural network.
	\item MTGNN~\cite{wu2020connecting}: A graph neural network based multivariate time series forecasting approach.
	\item MTGNN+sampling~\cite{wu2020connecting}: The MTGNN model trained on a sampled subset in each iteration.
\end{itemize}
\paragraph{Evaluation Metrics}
\label{subsec:timeseires_metric}
Following~\cite{lai2018modeling,shih2019temporal,wu2020connecting}, we use the Root Relative Squared Error (RRSE) and Empirical Correlation Coefficient (CORR) as metrics to evaluate the forecasting results.
The RRSE is a scaled version of the widely used Root Mean Square Error (RMSE), designed to make a more readable evaluation, regardless of the data scale.
The lower RRSE value and higher CORR value indicate better performance.

\begin{table}[t]
	\centering
	\begin{threeparttable}
		\resizebox{0.474\textwidth}{!}{
			\begin{tabular}{l| c c| c c c c |c c c c}
				\toprule
				\multirow{3}{*}{Dataset} & \multirow{3}{*}{$ l $} & \multirow{3}{*}{$\gamma$}& \multicolumn{4}{c|}{$k$} &   \multicolumn{4}{c}{$N$} \\
				\cmidrule(lr){4-11}
				&  & & \multicolumn{4}{c|}{Horizon} &   \multicolumn{4}{c}{Horizon} \\
				\cmidrule(lr){4-11}
				& & & 3& 6&12&24& 3& 6&12&24\\
				\midrule
				\midrule
				Traffic  &256   & 0.0001 &30  &5&10 &3 & 20 &30 & 30 &10 \\
				Electricity&512 & 0.0001 &5  &3&10  &5&20&3& 5& 20 \\
				Exchange-Rate &512 &0.0001 &20  &5&10  &5&20  &10&10  &20 \\
				\bottomrule
			\end{tabular}
		}
	\end{threeparttable}
	\caption{The selections of the hyper-parameters.}
	\label{table:alpha}
\end{table}

\begin{table*}
	\centering
	\begin{threeparttable}
		\resizebox{\textwidth}{!}{
			\begin{tabular}{lr|cccc|cccc|cccc}			
				\toprule
				\multicolumn{2}{c|}{Dataset}  & \multicolumn{4}{c|}{Traffic} & \multicolumn{4}{c|}{Electricity} & \multicolumn{4}{c}{Exchange-Rate}\\
				\midrule
				\multirow{2}{*}{Methods} &\multirow{2}{*}{Metrics } & \multicolumn{4}{c|}{Horizon} & \multicolumn{4}{c|}{Horizon} & \multicolumn{4}{c}{Horizon}  \\
				\cmidrule(lr){3-14}
				
				&  & 3 & 6 & 12 & 24 & 3 & 6 & 12 & 24 & 3 & 6 & 12 & 24 \\
				\midrule
				\midrule
				\multirow{2}{*}{AR}&RRSE ($\downarrow$)&0.5991&0.6218&0.6252&0.63&0.0995&0.1035&0.1050&0.1054&0.0228&0.0279&0.0353&0.0445\\
				&CORR ($\uparrow$)&0.7752&0.7568&0.7544&0.7519&0.8845&0.8632&0.8591&0.8595&0.9734&0.9656&0.9526&0.9357\\
				\midrule
				\multirow{2}{*}{VARMLP}&RRSE ($\downarrow$)&0.5582&0.6579&0.6023&0.6146&0.1393&0.1620&0.1557&0.1274&0.0265&0.0394&0.0407&0.0578\\
				&CORR ($\uparrow$)&0.8245&0.7695&0.7929&0.7891&0.8708&0.8389&0.8192&0.8679&0.8609&0.8725&0.8280&0.7675\\			
				\midrule				
				\multirow{2}{*}{GP}&RRSE ($\downarrow$)&0.6082&0.6772&0.6406&0.5995&0.1500&0.1907&0.1621&0.1273&0.0239&0.0272&0.0394&0.0580\\
				&CORR ($\uparrow$)&0.7831&0.7406&0.7671&0.7909&0.8670&0.8334&0.8394&0.8818&0.8713&0.8193&0.8484&0.8278\\
				\midrule			
				\multirow{2}{*}{RNN-GRU}&RRSE ($\downarrow$)&0.5358&0.5522&0.5562&0.5633&0.1102&0.1144&0.1183&0.1295&0.0192&0.0264&0.0408&0.0626\\
				&CORR ($\uparrow$)&0.8511&0.8405&0.8345&0.8300&0.8597&0.8623&0.8472&0.8651&0.9786&\underline{0.9712}&0.9531&0.9223\\
				\midrule
				\multirow{2}{*}{LSTNet-skip}	& RRSE ($\downarrow$) 	& 0.4777 & 0.4893 & 0.4950 & 0.4973	& 0.0864 & 0.0931 & 0.1007 & 0.1007 & 0.0226 & 0.0280 & 0.0356 & 0.0449\\
				& CORR ($\uparrow$) & 0.8721 & 0.8690 & 0.8614 & 0.8588 & 0.9283 & 0.9135 & 0.9077 & 0.9119 & 0.9735 & 0.9658 & 0.9511 & 0.9354\\
				\midrule
				\multirow{2}{*}{TPA-LSTM}	& RRSE ($\downarrow$) 	& 0.4487 & 0.4658 & 0.4641 & 0.4765	& 0.0823 & 0.0916 & 0.0964 & 0.1006	&\underline{0.0174} & \underline{0.0241} & \underline{0.0341} & \underline{0.0444}\\
				& CORR ($\uparrow$)  & 0.8812 & 0.8717 & 0.8717 & 0.8629 & 0.9439 & 0.9337 & 0.9250 & 0.9133 & \underline{0.9790} & 0.9709 & 0.9564 & 0.9381\\
				\midrule			
				\multirow{2}{*}{MTGNN} &RRSE ($\downarrow$)  & \underline{0.4162}  & 0.4754 & \underline{0.4461} & \underline{0.4535} & \underline{0.0745} & 0.0878 & \underline{\textbf{0.0916}} & \underline{\textbf{0.0953}} & 0.0194 & 0.0259 & 0.0349 & 0.0456 \\
				& CORR ($\uparrow$)   & \underline{0.8963} & 0.8667 & \underline{0.8794} & \underline{\textbf{0.8810}} & \underline{0.9474} & 0.9316 & \underline{0.9278} & \underline{0.9234} & 0.9786 & 0.9708 & 0.9551 & 0.9372 \\	
				\midrule
				MTGNN &RRSE ($\downarrow$) &0.4170& \underline{0.4435}& 0.4469&0.4537&0.0762& \underline{0.0862} &0.0938&0.0976&0.0212&0.0271&0.0350&0.0454\\
				+sampling&CORR ($\uparrow$)  & 0.8960& \underline{0.8815} & 0.8793&0.8758&0.9467& \underline{0.9354} &0.9261&0.9219&0.9788&0.9704&\underline{0.9574}&\underline{0.9382}\\ 
				%				\multirow{2}{*}{HyDCNN} &RRSE ($\downarrow$)  &0.4198&\underline{0.4290}&0.4352&0.4423&0.0832&0.0898&0.0921
				%			&\underline{0.0940}&-&-&-&-\\
				%				&CORR ($\uparrow$) & 0.8915&\underline{0.8855}&0.8858&0.8819&0.9354&0.9329&\underline{0.9285}&\underline{0.9264}&-&-&-&-\\ 
				\midrule
				\midrule
				\multirow{2}{*}{\textbf{IGMTF}} &RRSE ($\downarrow$)&\textbf{0.4135}&\textbf{0.4319}&\textbf{0.4422}&\textbf{0.4471}&\textbf{0.0740}&\textbf{0.0851}&0.0930&0.0983&\textbf{0.0173}&\textbf{0.0239}&\textbf{0.0329}&\textbf{0.0427}\\
				&CORR ($\uparrow$) &\textbf{0.8978}&\textbf{0.8879}&\textbf{0.8825}&0.8796&\textbf{0.9508}&\textbf{0.9404}&\textbf{0.9311}&\textbf{0.9240}&\textbf{0.9796}&\textbf{0.9718}&\textbf{0.9577}&\textbf{0.9391}\\ 
				\bottomrule
			\end{tabular}
		}
	\end{threeparttable}
	\caption{Comparison of multivariate time series forecasting methods. The results of baselines are reported in~\cite{wu2020connecting}. The results with underline indicate the best results in compared methods; the results in bold are the best results on all methods.}
	\label{table:single}
\end{table*}

\begin{table*}[t]
	\centering
	\begin{threeparttable}
		\resizebox{0.74\textwidth}{!}{
			\begin{tabular}{cl|cccc|cccc}			
				\toprule
				\multicolumn{2}{c|}{Dataset}  & \multicolumn{4}{c|}{Traffic} & \multicolumn{4}{c}{Electricity}\\
				\midrule
				\multirow{2}{*}{Metrics}&	\multirow{2}{*}{Methods}  & \multicolumn{4}{c|}{Horizon} & \multicolumn{4}{c}{Horizon}   \\
				\cmidrule(lr){3-10}
				
				&  & 3 & 6 & 12 & 24 & 3 & 6 & 12 & 24 \\
				\midrule
				\multirow{3}{*}{RRSE ($\downarrow$)} &IGMTF\_NS&0.4172&0.4591&0.4466&0.4515&0.0756&0.0863&0.0955&0.0994\\
				& IGMTF\_NW&0.4185&0.4366&0.4456&0.4512&0.0746&0.0865&0.0943&0.1005\\
				& IGMTF &\textbf{0.4135}&\textbf{0.4319}&\textbf{0.4422}&\textbf{0.4471}&\textbf{0.0740}&\textbf{0.0851}&\textbf{0.0930}&\textbf{0.0983}\\
				\midrule 
				\midrule
				\multirow{3}{*}{CORR ($\uparrow$)} & IGMTF\_NS &0.8963&0.8757&0.8797&0.8766&0.9494&0.9384&0.9295&0.9234\\
				& IGMTF\_NW &0.8958&0.8862&0.8808&0.8772&0.9501&0.9390&0.9301&0.9229\\
				& IGMTF &\textbf{0.8978}&\textbf{0.8879}&\textbf{0.8825}&\textbf{0.8796}&\textbf{0.9508}&\textbf{0.9404}&\textbf{0.9311}&\textbf{0.9240}\\ 
				\bottomrule
			\end{tabular}
		}
		%traffic 24, IGMTF_NS, 0.4500, 0.8775
	\end{threeparttable}
	\caption{Comparison of IGMTF\_NS, IGMTF\_NW, and IGMTF on Traffic and Electricity datasets.}
	\label{table:ablation}
\end{table*}
\paragraph{Implementation Details}
We implement our framework base on the PyTorch library~\cite{paszke2019pytorch}\footnote{Our source code and data are available at this repository: https://github.com/Wentao-Xu/IGMTF.}, and run on all experiments with a single NVIDIA Tesla V100 GPU.
We tune our framework using the grid search to select the optimal hyper-parameters based on the performance of the validation set.
We search the number of hidden units $l$ in GRU and MLP in \{128, 256, 512, 1024\}; the number of closest time stamps $k$ (in training instance sampler) and the number of closest neighbors $N$ (in graph aggregation module) in \{3, 5, 10, 20, 30\}; the learning rate $\gamma$ in \{0.002, 0.001, 0.0005, 0.0001, 0.00005\}.
Table~\ref{table:alpha} list the best selections of hyper-parameters on different datasets and horizons. 
Besides, as point out in Section~\ref{subsec:mini-batch-encoder}, the batch size of our framework is the number of instances on each time stamp, and the number of training epoch is $100$ for all datasets.

\subsection{Main Results}
Table~\ref{table:single} shows the results of RRSE and CORR on Traffic, Electricity, and Exchange-Rate datasets. 
On the RRSE metrics, our framework achieves the best results on the Electricity dataset when the horizon is 3 and 6, and the best results on Traffic and Exchange-Rate in all horizons. 
On the CORR metrics, our frame outperforms the compared methods except for the Traffic dataset when the horizon is 24. 
Although on some metrics in some horizons, our IGMTF framework is worse than MTGNN or MTGNN+sampling, the IGMTF still performs better than the rest of the baselines.

The results in Table~\ref{table:single} verify the effectiveness of our IGMTF framework. In most cases, our framework is better than MTGNN, which utilizes the graph neural network to mine the inter-dependencies among variables in the same time stamp. 
Compared with MTGNN, our IGMTF can utilize the correlations between different variables in different time stamps.
Therefore, capturing the interdependencies between different variables in different time stamps can further improve multivariate time series forecasting performance.

\subsection{Ablation Study}
\label{subsec:ablation}
We apply an ablation study to study the effect of some components in our framework, including the training instance sampler in Section~\ref{subsec:sampler} and the mapping matrices $W_h$ and $W_e$ of graph aggregation module in Section~\ref{subsec:graph}.
To study the effect of these two components, we compare our IGMTF framework with the following  IGMTF' variants:
\begin{itemize}
	\item  IGMTF\_NS: To study the effect of the training instance sampler in Section~\ref{subsec:sampler}, the IGMTF\_NS remove the training instance sampler and randomly sample the same number of instances as sampled training instances.
	\item  IGMTF\_NW: To study the effect of the mapping matrices $W_h$ and $W_e$ of graph aggregation module in Section~\ref{subsec:graph}, the IGMTF\_NW remove $W_h$ and $W_e$ from the IGMTF framework.
\end{itemize}

Table~\ref{table:ablation} shows the results of  IGMTF\_NS, IGMTF\_NW, and IGMTF at Traffic and Electricity datasets. 
From Table~\ref{table:ablation} we can find that removing the training instance sampler or matrices $W_h$ and $W_e$ of the graph aggregation module would both reducing the performance of the IGMTF framework. 
Specifically, without a training instance sampler, our IGMTF framework would significantly weaken the forecasting results on the Traffic dataset with a horizon of 6.
These results illustrate the training instance sampler is vital for our IGMTF framework to sample the most related instances for mini-batch instances from all training instances.
The ablation study results also verify that the mapping matrices in the graph aggregation module are necessary and can improve the performance of our framework.

\subsection{Hyper-parameters Analysis}
In the training instance sampler, we have a hyper-parameter $k$ to control the number of time stamps we sample from the training instances. 
Meanwhile, there is also a hyper-parameter $N$ in the graph aggregation module to select the number of neighbors to aggregate information from sampled training instances to each mini-batch instance.
In this subsection, we will study the influence of hyper-parameters $k$ and $N$ on the performance of our IGMTF framework.

\begin{figure}[t]
	\centering
	\includegraphics[width=1\columnwidth]{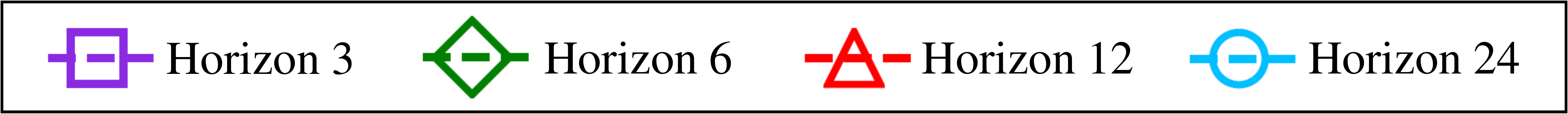}
	
	\subfigure[RRSE on Traffic.]{\includegraphics[width=0.47\columnwidth]{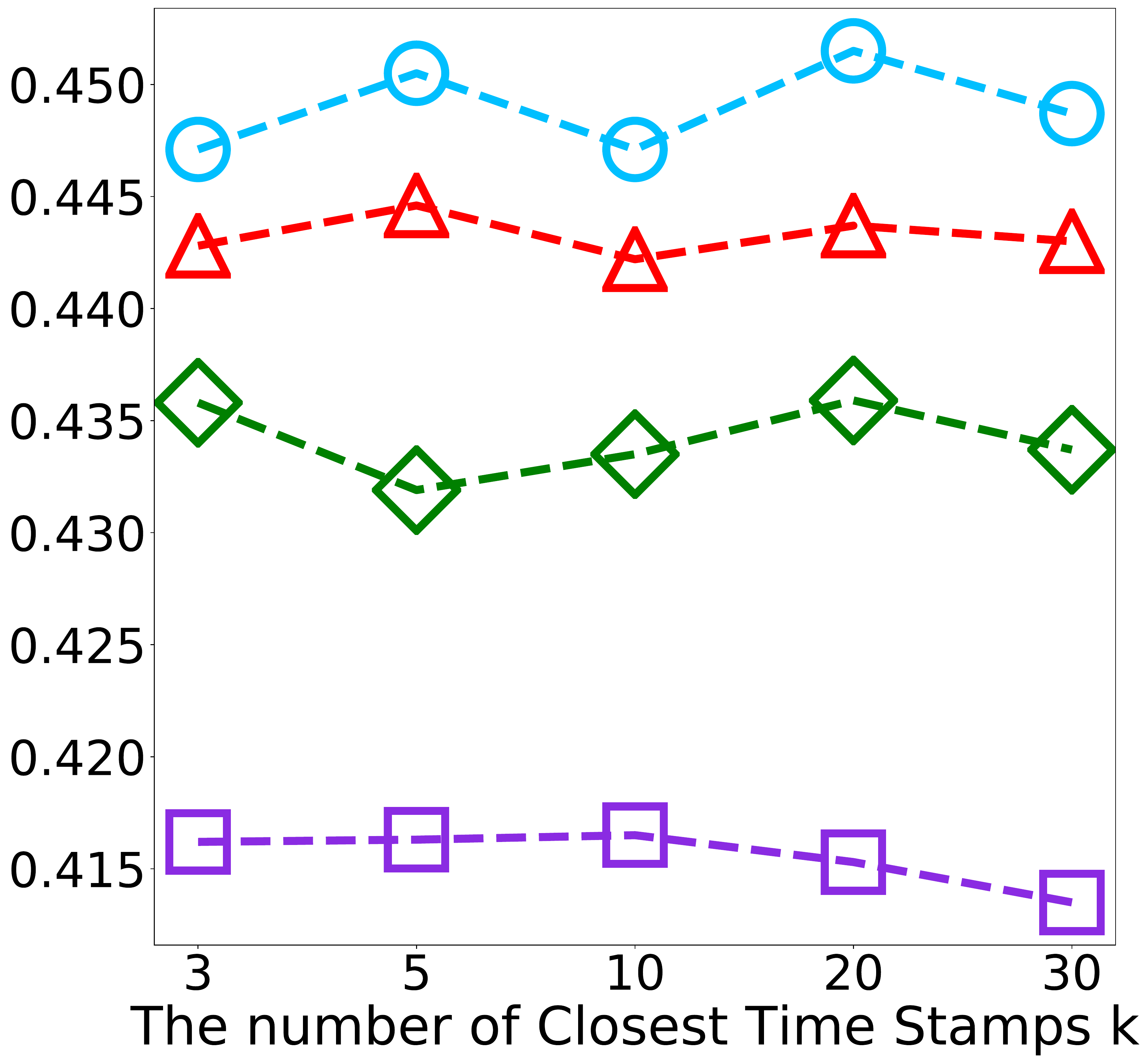}}
	\hspace{.1in}
	\subfigure[RRSE on Electricity.]{\includegraphics[width=0.47\columnwidth]{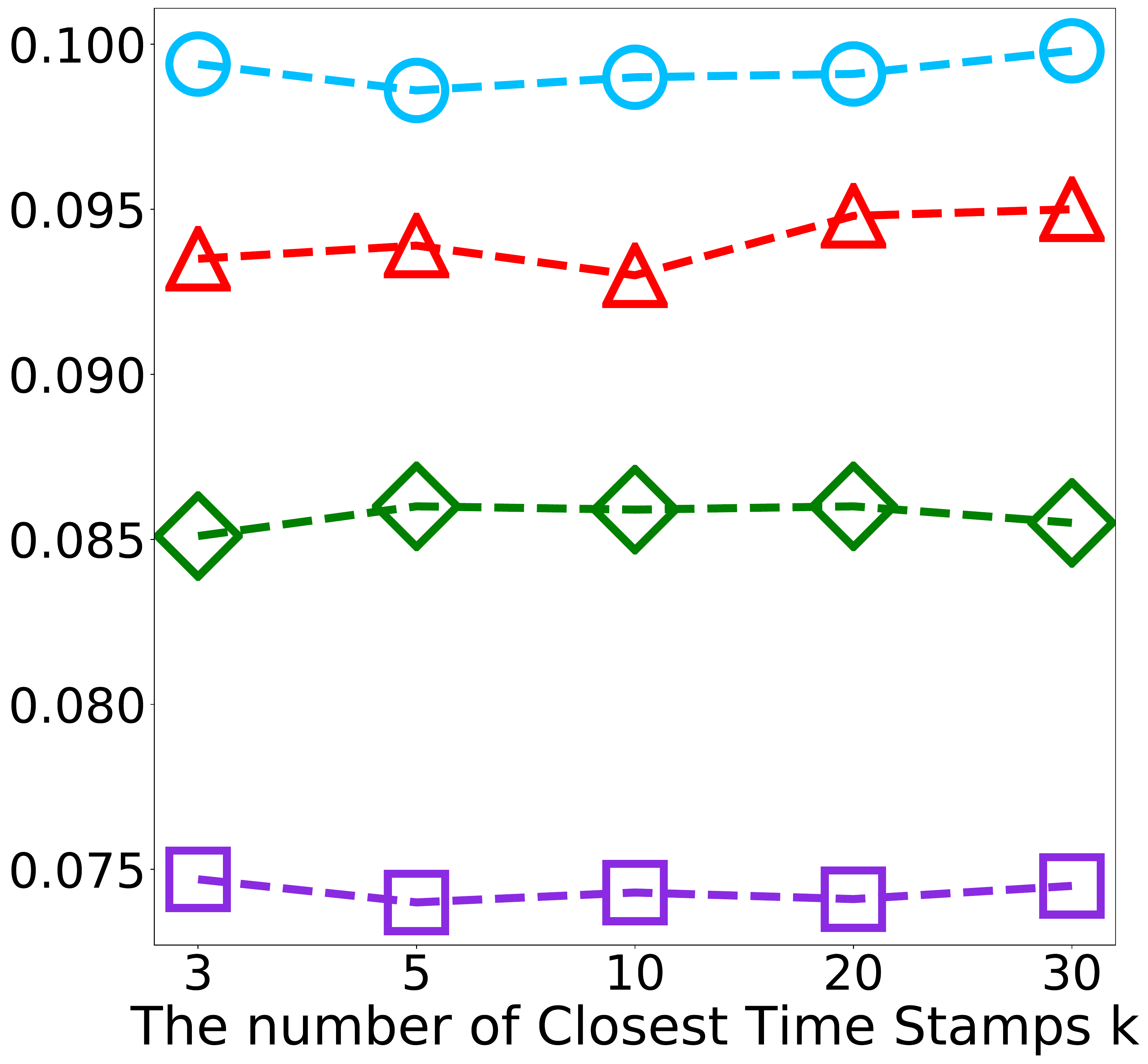}}
	\subfigure[CORR on Traffic.]{\includegraphics[width=0.47\columnwidth]{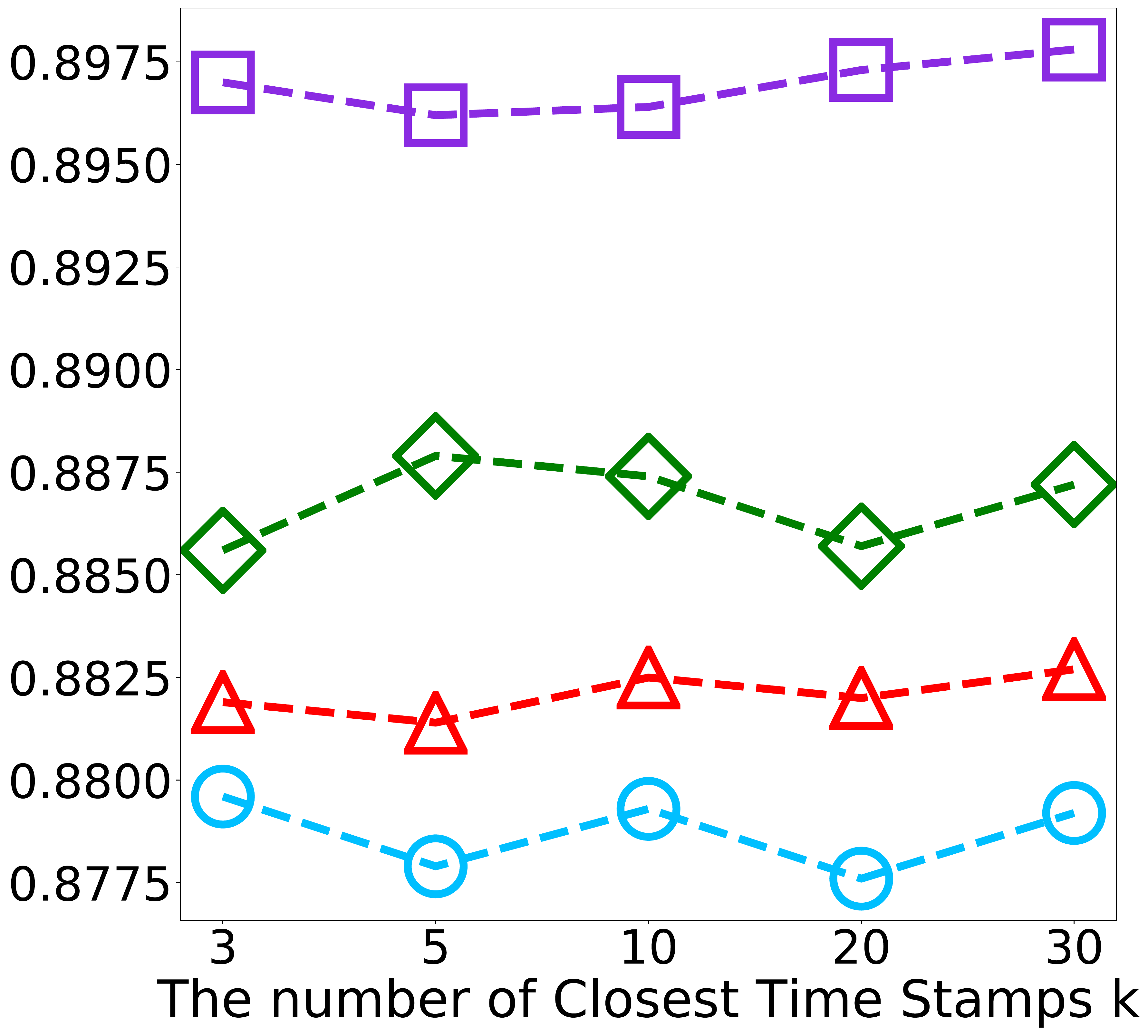}}
	\hspace{.1in}
	\subfigure[CORR on Electricity.]{\includegraphics[width=0.47\columnwidth]{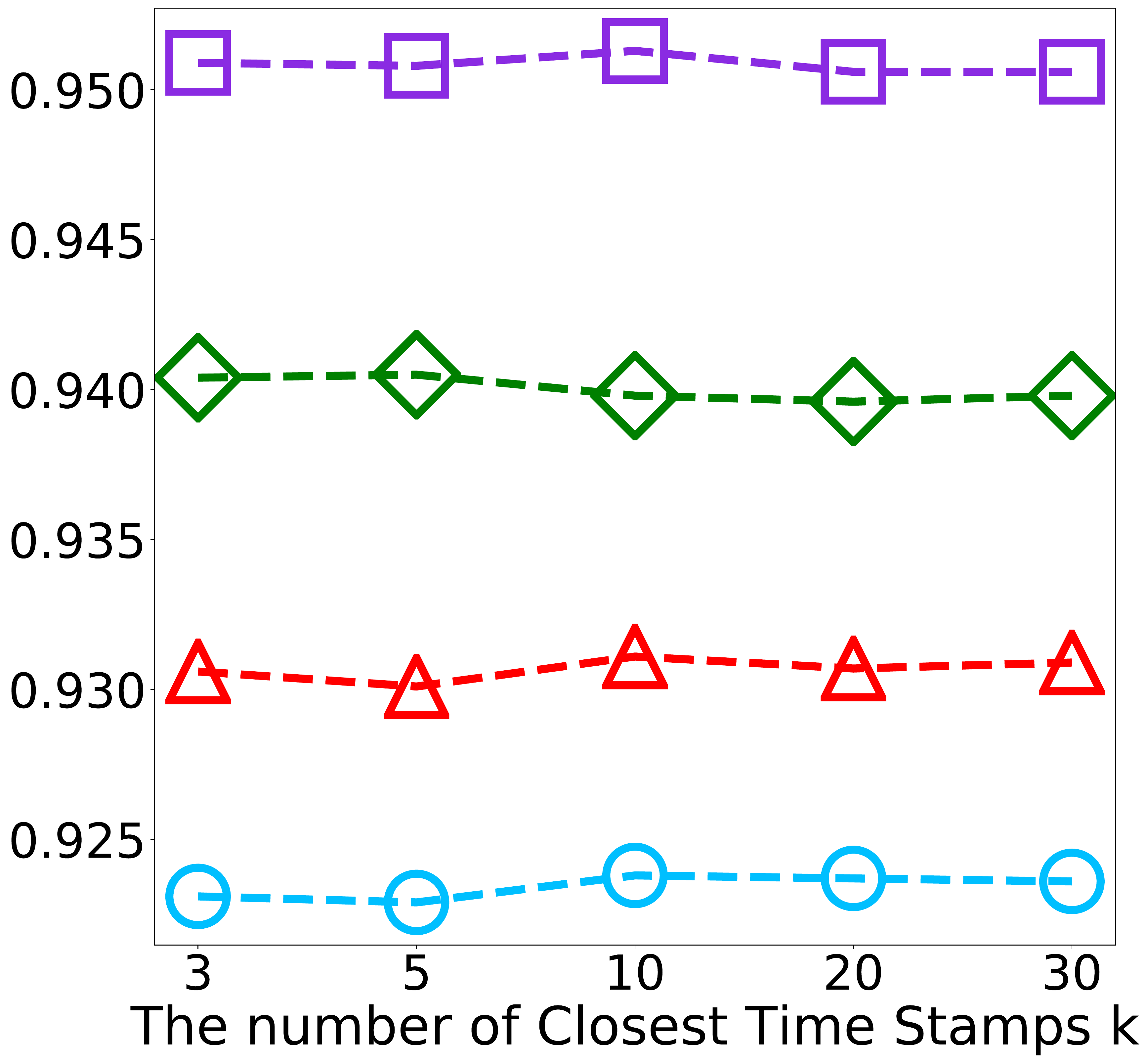}}
	\caption{RRSE and CORR results on Traffic and Electricity datasets with different numbers of closest time stamps $k$.}
	\label{fig:effect_k}
\end{figure}

\paragraph{Influence of the Number of Closest Time Stamps $k$}
We let $k$ vary in \{3, 5, 10, 20, 30\}, and we set the hyper-parameter $N$ as 10 and fix all the other hyper-parameters. 
Then we observe the RRSE and CORR results on Traffic and Electricity datasets under different $k$.
Figure~\ref{fig:effect_k} shows the results under different $k$. 
The influence of $k$ is more significant on the Traffic dataset than the Electricity dataset. 
On the Traffic dataset, IGMTF achieves the best results at horizons 3, 6, 12, 24 when $k$ is 30, 5, 10, 3, respectively.  
On the Electricity dataset, IGMTF achieves the best results at horizons 3, 6, 12, 24 when $k$ is 5, 3, 10, 5, respectively. 

%\begin{figure}[t]
%	\centering
%	\includegraphics[width=1\columnwidth]{figure/legend1}
%\end{figure} 

\begin{figure}[t]
	\centering
	\includegraphics[width=1\columnwidth]{figure/legend1}

	\subfigure[RRSE on Traffic.]{\includegraphics[width=0.47\columnwidth]{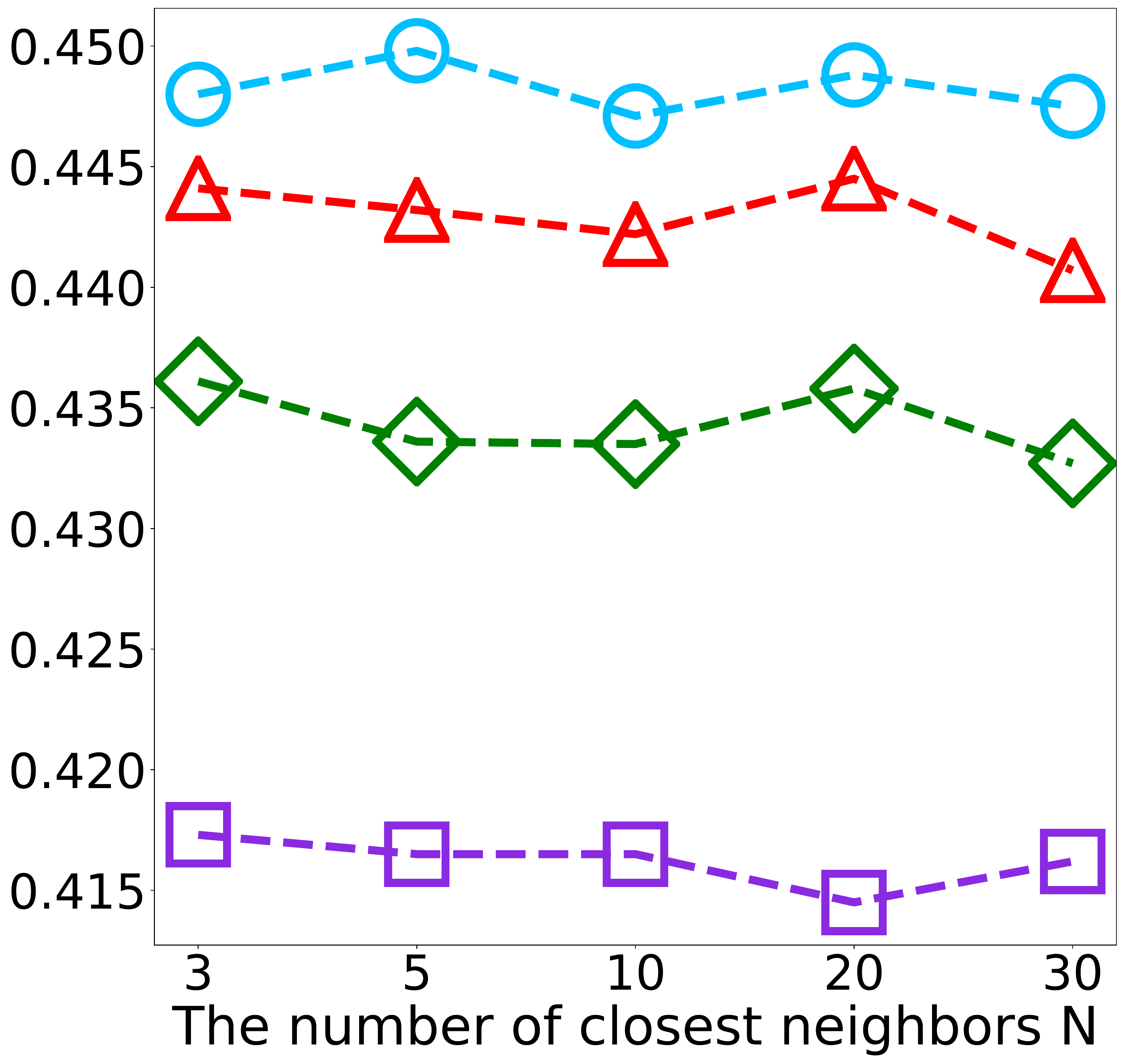}}
	\hspace{.1in}
	\subfigure[RRSE on Electricity.]{\includegraphics[width=0.47\columnwidth]{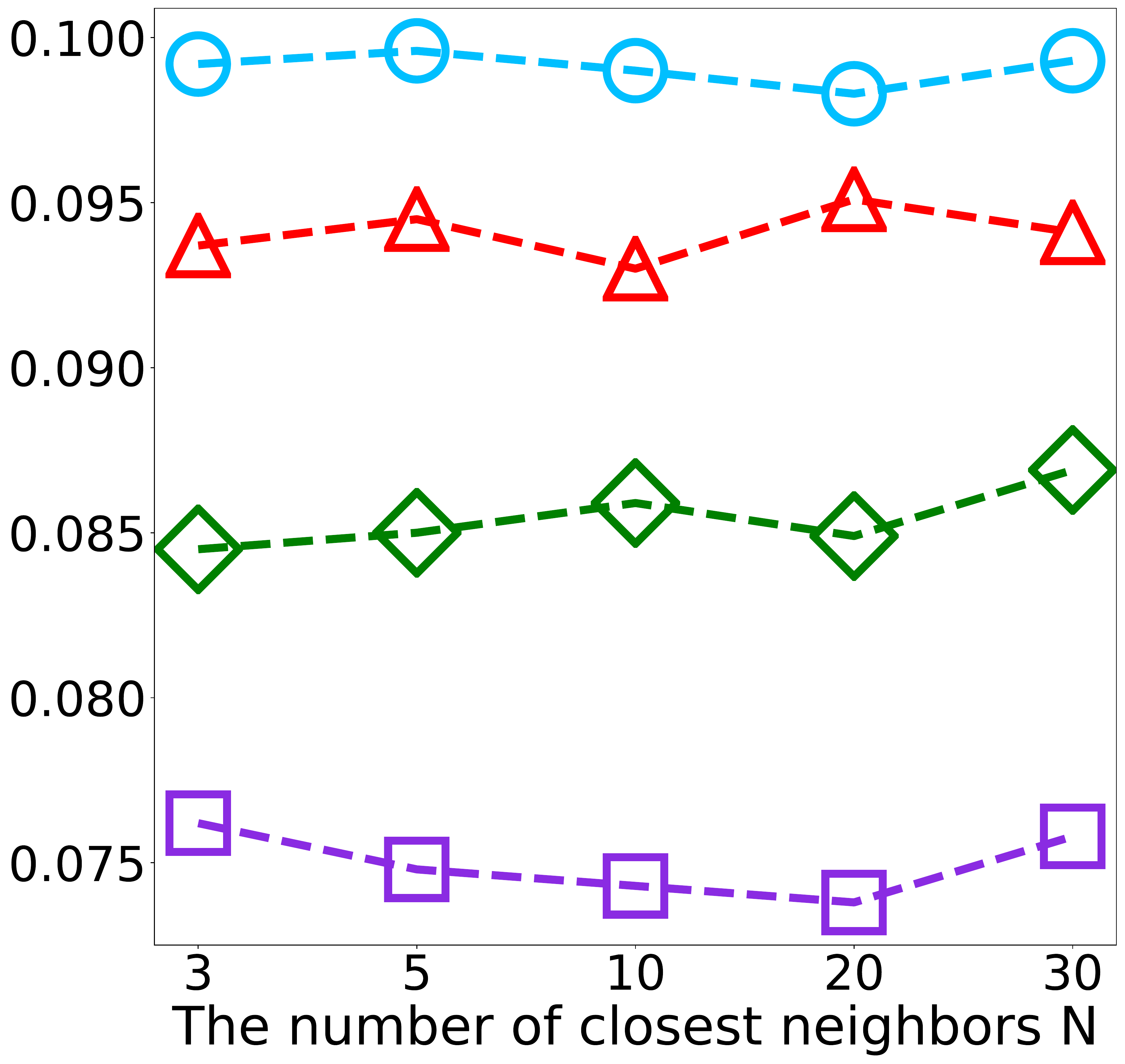}}
	\subfigure[CORR on Traffic.]{\includegraphics[width=0.47\columnwidth]{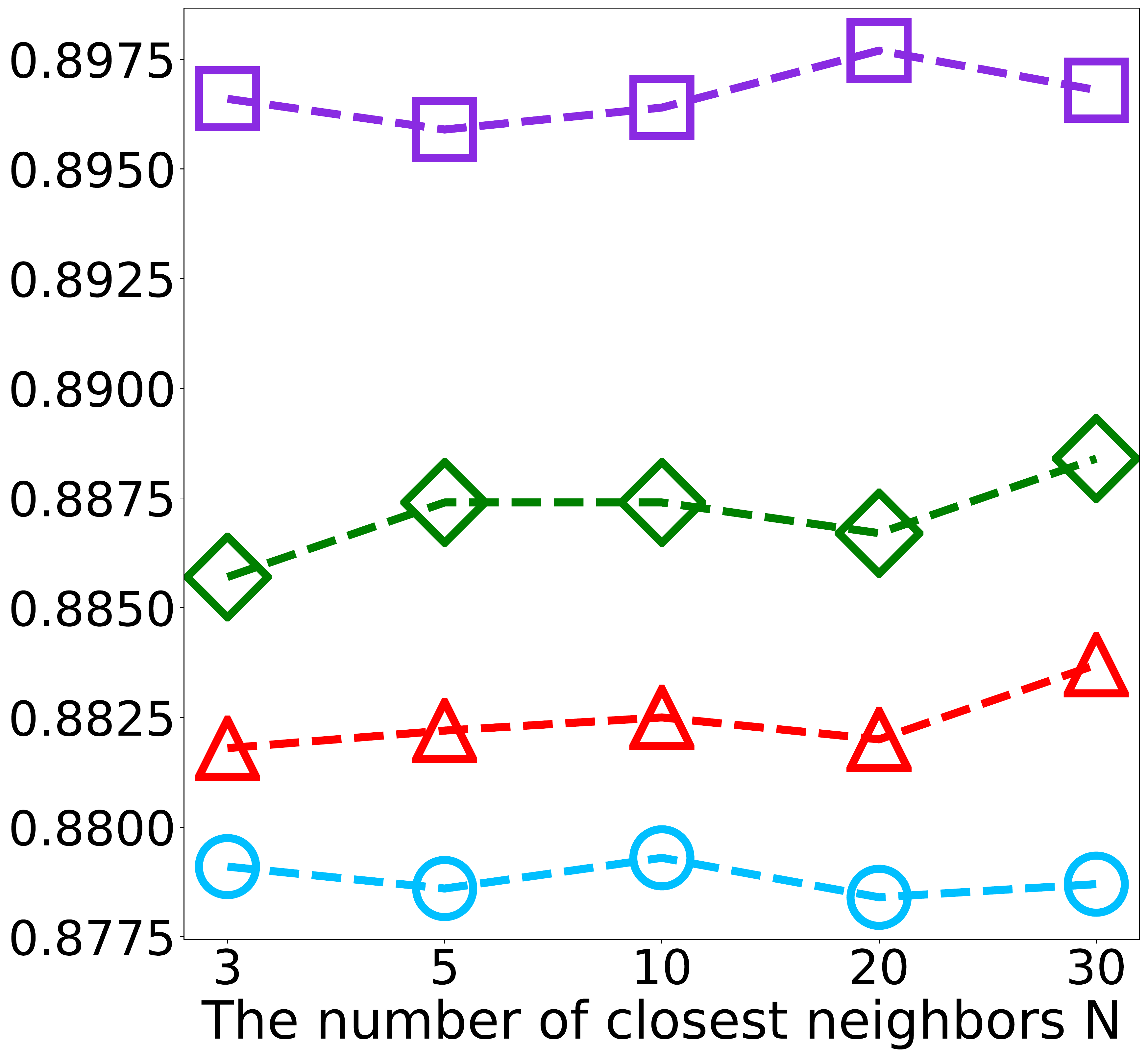}}
	\hspace{.1in}
	\subfigure[CORR on Electricity.]{\includegraphics[width=0.47\columnwidth]{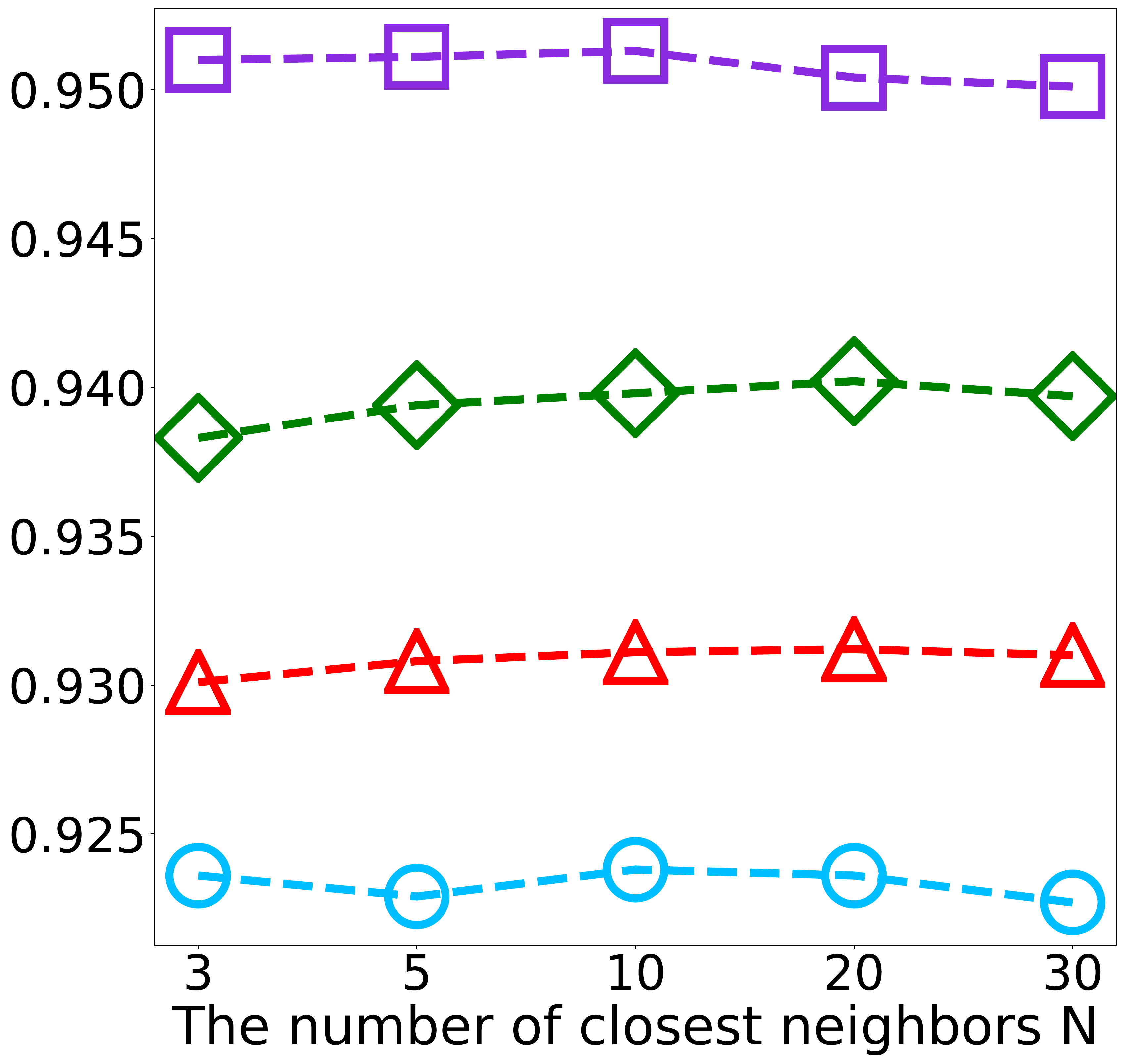}}
	\caption{RRSE and CORR results on Traffic and Electricity datasets with different numbers of closest neighbors $N$.}
	\label{fig:effect_n}
\end{figure} 
\paragraph{Influence of the Number of Closest Neighbors $N$}
We let $N$ vary in \{3, 5, 10, 20, 30\}, and we set the hyper-parameter $k$ as 10 and fix all the other hyper-parameters. 
Then we observe the RRSE and CORR results on Traffic and Electricity datasets under different $N$.
Figure~\ref{fig:effect_n} shows the results under different $N$.
On the Traffic dataset, IGMTF achieves the best results at horizons 3, 6, 12, 24 when $N$ is 20, 30, 30, 10, respectively.  
On the Electricity dataset, IGMTF achieves the best results at horizons 3, 6, 12, 24 when $N$ is 20, 3, 5, 20, respectively. 

The influence of the numbers of closest time stamps $k$ and neighbors $N$ suggest that both the $k$ and $N$ are sensitive hyper-parameters that need to be tuned for the best performance given a dataset and a horizon.

\section{Conclusion and Future Work}
\label{sec:conclusion}
In this work, we propose a simple yet efficient instance-wise graph-based framework (IGMTF) for multivariate time series forecasting. 
Our framework can address exiting work's limitation that overlooks the interdependencies between different variables at different time stamps.
The key idea of our framework to address this limitation is aggregating information from historical training instances to mini-batch instances.
We evaluate our framework on the multivariate time series benchmark datasets.
The experimental results show that our proposed model performs better than the state-of-the-art baseline methods. 

In the future, we plan the explore more techniques, such as contrastive learning, to mine valuable interdependences between different time series at different time stamps.

\bibliographystyle{aaai22}
\bibliography{aaai22}

\end{document}